# Play MNIST For Me!
# User Studies on the Effects of Post-Hoc, Example-Based Explanations & Error Rates on Debugging a Deep Learning, Black-Box Classifier

**Courtney Ford**[1], **Eoin M. Kenny,**[1,2,3] **& Mark T. Keane**[1,2,3]

[1]School of Computer Science, University College Dublin, Dublin, Ireland
[2]Insight Centre for Data Analytics, UCD, Dublin, Ireland
[3]VistaMilk SFI Research Centre, Ireland

## Abstract

This paper reports two experiments (N=349) on the impact of post-hoc explanations-by-example and error-rates on people's perceptions of a black-box classifier. Both experiments show that when people are given case-based explanations, from an implemented ANN-CBR twin system, they perceive miss-classifications to be more correct. They also show that as error rates increase above ~4%, people trust the classifier less and view it as being less correct, less reasonable and less trustworthy. The implications of these results for XAI are discussed.

## 1 Introduction

The recent explosion of research on Explainable Artificial Intelligence (XAI) has been marked by the galloping development of computational techniques trailed by a flagging walk of user studies [Miller *et al.*, 2017; Mueller *et al.*, 2019]. This raises the prospect that many XAI models may fail psychological validation with end-users, undermining the realisation of XAI. One notable exception to this trend is the DARPA XAI program [Gunning, 2017; Gunning and Aha, 2019; Mueller *et al.*, 2019] which is testing end-users in lock-step with technique development (e.g., see [Glickenhaus *et al.*, 2019]); notably, several groups have in this program have evaluated XAI techniques for black-box AI models, such as Convolutional Neural Networks (CNNs) using image datasets (see section 5). Here, we also try to redress this user-testing deficit by evaluating post-hoc, case-based explanations of a CNN's classifications, involving the MNIST dataset from a twin systems XAI perspective; twin systems explain black-box artificial neural networks by mapping their feature-weights to case-based reasoner [Kenny and Keane, 2019; Keane and Kenny, 2019b].

Many user studies have evaluated XAI for AI classifiers by requiring people to debug the machine learning system, in contexts where people have high levels of expertise (e.g., reading hand-written numbers; Bäuerle, *et al.* [2018]). This debugging task is also used here, as people are asked to make correctness and reasonableness judgements about right and wrong classifications made by a CNN on the written digits of the MNIST dataset [LeCun et al., 1998]. Two key issues are examined: namely, how people's perception of this black-box classifier is impacted by (i) post-hoc explanations-by-example (ii) error rates in the classifier's performance. In the remaining sections of this introduction, we sketch the broader context for this work and some specific prior work that has recently been carried out (see also section 5).

### 1.1 Positioning the Current Work

Echoing distinctions in Philosophy and Psychology between *explanation* and *justification*, Lipton's [2018] analysis of *interpretability* distinguishes between *transparency* (i.e., "*How does the model work*") and *post hoc explanation* (i.e., "*What else can the model tell me?*"), though he admits that one may shade into the other. The current work tests post-hoc explanation-by-example or case-based explanation [Sørmo et al., 2005; Nugent and Cunningham, 2005]. It has long been argued in case-based reasoning (CBR; [Aamodt and Plaza, 1994; Mantaras *et al.,* 2006]) that cases/exemplars provide plausible and intuitive explanations for humans [Leake, 1996; Leake and McSherry, 2005; Keane and Kenny, 2019a, 2019b]. However, the empirical testing of this proposal has often been found wanting; Keane and Kenny's [2019a] review found that ~1% of CBR papers on explanation carried out adequate user testing.

Doshi-Velez and Kim [2017] have distinguished a three-leveled taxonomy for XAI evaluations: (i) *application grounded evaluation*, where human end-users (typically with some domain expertise) are tested using the complete AI model in the task for which it was built; (ii) *human-grounded evaluation*, where human end-users (who may not be domain experts) test selective aspects of the "real" application task with the AI model; (iii) *functionally-grounded evaluation*, where the evaluation is based on some computational-proxy for an actual human evaluation (i.e., no human testers). Hence, end-user responsibility is reduced across the three levels from end-users evaluating real explanations generated by the complete system; to end-users responding to general aspects of explanation (or localized system performance) to no user involvement at all. The present work mainly belongs to the first level of this taxonomy (i.e., we have human experts evaluating the outputs of the AI model), though some aspects of it shade into the second level (i.e., people are not debugging a running system, as in [Bäuerle *et al.,* 2018]). Rather, the people in our study are performing a judgment task, rating *actual* outputs from the system on several dimensions (i.e., correctness, reasonableness, trust, satisfaction), as indirect measures of an explanation's impact.

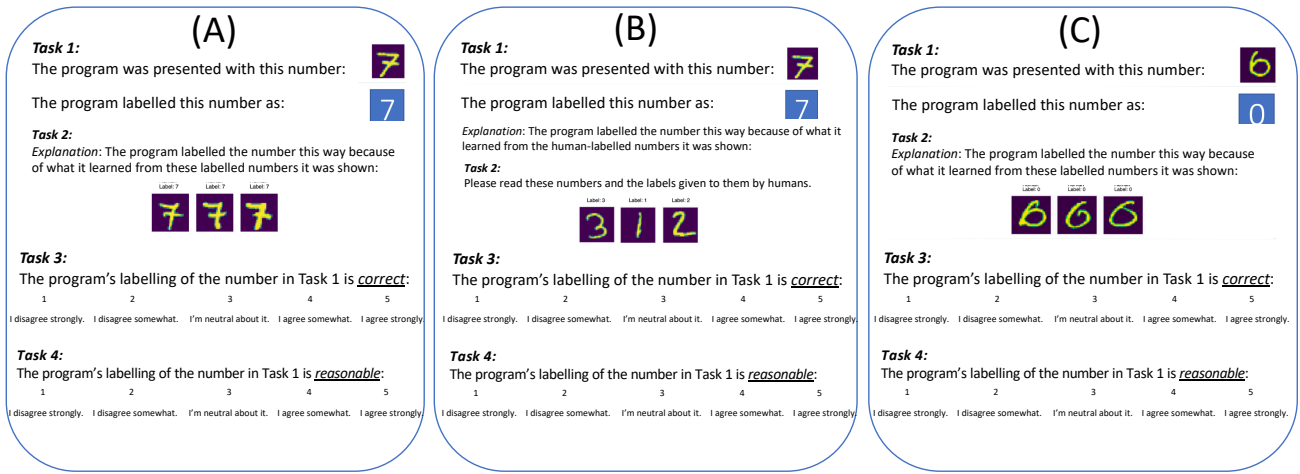

*Figure 1. Sample materials from KQK [2020] showing (A) a right classification of an MNIST query-item with a case-based explanation present and (B) absent, along with the correctness and reasonableness rating-tasks, (C) a wrong classification (where the model classifies a "6" as a "0", relying on training data with "0s" that look like "6s") with its case-based explanation.*

Finally, recently, Kenny, Quinn and Keane [2020] proposed a distinction between studies doing local or global tests of explanatory validity. *Local tests* concentrate on the impact of explanations of an AI system at the level of single predictions. *Global tests* concentrate on the impact of explanations on people's overall perception of and performance of an AI system; global tests typically use performance measures (e.g., are people faster using the system) or subjective judgements (e.g., people's trust/satisfaction in the system; see [Hoffman *et al.*, 2018]). The present experiments perform indirect, local tests of the impact of explanations using correctness or reasonableness ratings, as well as global tests based on the DARPA trust/satisfaction surveys. As we shall see, this local-global distinction raises a number of important issues for XAI (see section 6). In the next section, we review the specific prior work that informs the current studies. (see section 5 for a wider review).

## 1.2 A Test of Post-Hoc Case-Based Explanations

The present work follows on from a user-test of the twin system approach (see Kenny, Quinn and Keane [2020]; henceforth, KQK). KQK explained a CNN's classifications involving MNIST images by using a feature-weighting method to analyze the CNN, to discover feature-weights that then are mapped to a *k*-NN search, operating over the same MNIST dataset, for explanatory cases. KQK's experiment presented people with right and wrong classifications from the CNN, asking them to judge the correctness and reasonableness of the classifications when post-hoc, case-based explanations were provided (see Fig. 1). In KQK's study, 80% of the classifications shown were right and 20% were wrong (note, the CNN's actual success rate was ~99%).

KQK found that explanations only impacted people's evaluations of the CNN's wrong classifications (as in Fig.1C), a finding also seen in the DARPA user-studies [Glickenhaus *et al.*, 2019]. Specifically, KQK found that when miss-classifications were explained using cases, people rated them as significantly more correct and sometimes more reasonable. KQK also used the DARPA trust/satisfaction surveys to measure people's global evaluation of the system and found that the explanations did not impact these measures. They concluded that, perhaps, people found the overall 20% error-rate to be unacceptable, irrespective of the explanations provided. In short, while the explanations made the errors locally more acceptable, they did not globally "explain away" the overall poor performance of the system; indeed, research on *algorithm aversion* shows that people have a very low bar for accepting any error in automated systems [Burton *et al.*, 2019; Dietvorst et al., 2015]. Clearly, KQK's work invites further exploration as to what error-rate might indeed be acceptable. This question is one that is common to all AI systems that provide users with predictions that may be occasionally wrong; here, we examine it in a debugging image-classifications task.

## 1.3 Outline of Current Paper

In the remainder of this paper, we report on the results of two experiments extending the KQK work. In Experiment 1, the local effects of providing explanations for the CNN's miss-classifications are tested using *high-end error-rates* of 30% and 60%, compared to a low baseline (of 3%-error; see sections 2 and 3). In Experiment 2, we again test the impact of case-explanations and *low-end error-rates* of 4%, 12%, 20%, and 28% (see sections 2 and 4). As we shall see, these studies which represent an advance on the current literature (see section 5), find that (i) case-based explanations locally influence people's correctness perceptions of the model's errors and (ii) error-rates impact local and global perceptions of the AI system (notably more in trust than in satisfaction). In section 6, we consider the broader implications of these findings for the interpretability of black box models.

## 2 Experiments 1 & 2: Method

As both experiments used roughly the same measures and material sets, the method from each is described here before reporting their different designs and results.

### 2.1 Participants

The two user studies involved people (N=349) recruited on the Prolific crowdsourcing site (www.prolific.com). All participants were aged over 18, native English speakers and lived in the USA, UK, or Ireland. Exclusion criteria were participation in previous studies by the lab and inattentive answering (4 people were excluded from Expt.1 for not noticing miss-classifications noted by all other participants). Participants were randomly assigned to the conditions of each experiment: Expt.1 (N=165), Expt.2 (N=184). These Ns were chosen based on a power analysis for a moderate effect-size.

### 2.2 Materials

Materials for both user studies were outputs from a CNN classifier using the MNIST dataset [Kenny and Keane, 2019]. The miss-classifications were actual errors produced by the model (i.e., query-items where the classification made differed from the ground truth). As the model produces correct classifications ~99% of time, these errors came from multiple runs. All errors were *alternate-labelling errors* in which the model gave a close but incorrect classification (see Fig.1C). Also, the explanatory cases used were the 3 nearest-neighbor images, from the MNIST training-set, found by the CNN-CBR twin system. Expt.1 used 30 distinct materials and Expt. 2 had 25 (see Fig.1 for examples).

### 2.3 Procedure, Measures & Analyses

After being told the system was a program that "learned" to classify written numbers, people were told they would be shown several examples of its classifications (see Fig. 1). Their task was to rate the correctness and reasonableness of the presented classification on a 5-point Likert-scale from "I disagree strongly" (1) to "I agree strongly" (5). In the Explanation-Present conditions, they were told that the model's classification was learned from human-labelled numbers (see Fig. 1A). In the Explanation-Absent conditions, they were given a visually-matched items, three unrelated number-images along with their labels (see Figure 1B) which were not explanatory cases. After rating all of the presented classifications, participants filled out the DARPA trust (8 questions) and satisfaction surveys (8 questions). In Expt.2, before doing these surveys, participants also rated the system's overall correctness and reasonableness on the presented items. So, the 4 measures were:

- *Correctness*. Mean 5-point Likert-scale ratings of correctness collated for right and wrong classifications.
- *Reasonableness*. Mean 5-point Likert-scale ratings of reasonableness collated for right & wrong classifications.
- *Global Correctness & Reasonableness*. (only in Expt.2) overall correctness & reasonableness ratings (5-pt scales).
- *Trust and Satisfaction*. Ratings from the DARPA Trust and Satisfaction surveys together and individually.

For both experiments, MANOVAs were computed for the independent variables (Explanation and Percent-Error) involving the dependent variables of (i) right/wrong classifications for correctness ratings, (ii) right/wrong classifications for reasonableness ratings, (iii) 8 trust-question ratings, and (iv) 8 satisfaction-question ratings.

## 3 Expt. 1: XPs & High-End Error-Rates

Experiment 1 (N=165) tested the effects of providing post-hoc, example-based explanations crossed with varying the relative percentage of error-items people saw in the set of classifications. The design was a 2 (Explanation: Present v Absent) x 3 (Percent-Error: 3% v 30%, v 60%) x 2 (Classification-Type: Right v Wrong) with Explanation and Percent-Error being between-participants variables and Classification-Type being a within-participant variable. Recall, in KQK's experiment people saw 30 items of which 24 were right (80%) and 6 that were wrong (20%).

### 3.1 Expt.1: Results & Discussion

When example-based explanations were present people perceive miss-classifications as being more correct, than when explanations are absent (replicating KKQ's findings, see Table 1 and Fig. 2); people rate wrong classifications as more correct, with an explanation, presumably because it shows the model working consistently but with miss-labelled data (see Fig.1C). Explanations also impact reasonableness ratings but less clearly. Increasing error-rates negatively impact people's ratings of correctness, reasonableness, and trust; notably, models with error-rates of 30%-60%, are trusted significantly less than ones with a 3% error-rate.

**3.1.1 Correctness.** The MANOVA analyses of correctness ratings revealed significant effects for Explanation and Percent-Error though the interaction was not statistically significant (see Table 1). However, all of these effects occur in people's ratings of wrong classifications, not in right classifications (see Fig. 2). Univariate analyses for the wrong classifications showed main effects for the Explanation, $F(1, 159) = 4.91$, $p < .05$, $\eta^2=0.03$, and Percent-Error variables, $F(2, 159) = 31.60$, $p < .001$, $\eta^2=0.28$. The analyses for right classifications show no significant effects (all $ps > .30$). Interestingly, people rate the wrong classifications as being more correct in the Explanation-Present ($M = 1.82$, $SD = 0.72$) than in the Explanation-Absent ($M = 1.62$, $SD = 0.58$) condition, a difference that is statistically significant, $p < .05$. In contrast, people's ratings of the right classifications are not reliably different (Explanation-Present, $M=4.73$, $SD=0.37$; Explanation-Absent, $M=4.67$, $SD=0.48$; see Fig. 2). Percent-Error also impacts correctness ratings for wrong classifications, $F(2, 159) = 31.61$ $p < .001$, $\eta^2 = .284$, but not right classifications, $F(2, 159) = 1.45$, $p > .20$, $\eta^2 = .018$. But, the effect is somewhat counter-intuitive; as people see more miss-classifications (30% or 60% versus 3%) they tend to rate the miss-classifications as being marginally more correct; 3% (*M* = 1.23, *SD* = 0.50), 30% (*M* = 1.93, *SD* = 0.54), 60% (*M* = 1.99, *SD* = 0.64). Post-hoc Tukey HSD comparisons for these wrong classifications show the differences between the 3%-30% and 3%-60% conditions to be reliable (*p*s <0.001; see Fig. 2). None of the pairwise comparisons for the right classifications were reliably different.

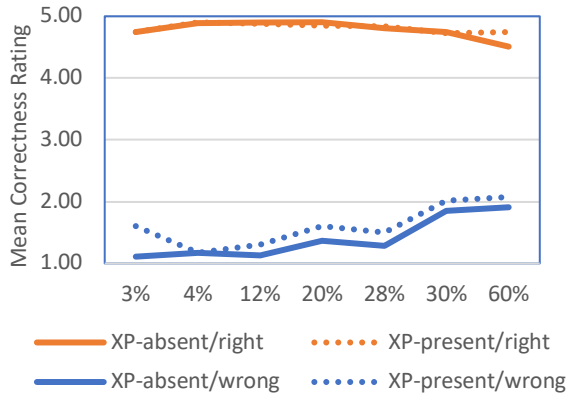

*Figure 2. Mean correctness ratings of right and wrong classifications when Explanations (XP) are present or absent for each Percent-Error level in Expts. 1 and 2*

**3.1.2 Reasonableness.** The pattern of results for reasonableness ratings is less clear. The MANOVA analyses of reasonableness ratings revealed a significant interaction between Explanation and Percent-Error, $F(4, 316) = 2.53$, $p < .05$, Wilks' $\Lambda = .94$, $\eta^2 = 0.03$. No main effects were statistically significant (see Table 1). Univariate analyses showed this Explanation x Precent-Error interaction occurs only for the right classifications, $F(2, 159) = 4.27$, $p < .05$, $\eta^2 = .05$; this interaction was specifically due to people rating the 60%-error condition as being more reasonable in the Explanation-present (*M* = 4.78, *SD* = .09) than in the Explanation-absent (*M* = 4.38, *SD* = .08) condition, a difference that is statistically reliable ($p < .05$). It is hard to interpret this finding. KKQ found that reasonableness was not consistently impacted by the provision of explanations (they were only found in one error-type). Our current view is that the reasonableness measure may not be as robust as correctness; it may suffer from different people having quite different interpretations of what reasonableness means.

**3.1.3 Trust.** The MANOVA analyses of the trust survey revealed a significant effect for Percent-Error, however, there was no main effect of Explanation or interaction (see Table 1). Univariate main effects for Percent-Error were found in all questions (all *p*s < .05), except for Question 7 ("The system can perform the task better than a novice human user"). Two interesting post-hoc Tukey HSD comparisons indicated that overall people in the 3%-error conditions reported the highest level of confidence in the system (*M* = 4.28, *SD* = .67), and rated the classifications as most predictable (*M* = 4.21, *SD* = .82). Also, users in the 3%-error condition enjoyed using the system for decision making (*M* = 3.11, *SD* = .96) significantly more than the people in the 60%-error one (*M* = 2.55, *SD* = 1.15). -So, in summary, trust is mainly impacted by the error-rates people encounter more than the provision of an explanation. The explanation seems to act locally affecting people's perception of the correctness of miss-classifications, but those explanations do not globally "explain away" the failures of the system.  Trust is impacted by rising error-rates though not linearly; from this study, it appears that trust levels decrease sharply at 30%-error rates (relative to 3%-errors) and then plateau to 60%-error.

**3.1.4 Satisfaction.** The MANOVA analyses of global satisfaction ratings of the overall system revealed a significant effect for Percent-Error (see Table 1). There was no main effect of Explanation or interaction. The satisfaction survey questions do not tell us much in this test context (similar results were found by KQK).  It is unclear whether this is an issue with the measures or this particular task.  Part of the problem here, may be that the satisfaction questions (unlike the trust ones) seem to range over a number of distinctly different issues and treating them as a unitary set is not sensible (Hoffman, 2020, personal communication).

## 4   Expt. 2: XPs & Low-End Error-Rates

Expt.1 tested the impacts of post-hoc explanations and error rates for *high-end error-rates* (30% and 60%, compared to 3%).  Experiment 2 (N=184) tested the Explanation and Percent-Error variables using *low-end error rates* to gain a fuller profile of error-rate effects. So, the design was a 2 (Explanation: Present v Absent) x 4 (Percent-Error: 4% v 12% v 20% v 28%) x 2 (Classification-Type: Right v Wrong) with Explanation and Percent-Error being between-participants variables and Classification-Type being a within-participant one. It also assessed overall correctness and reasonableness, after participants had seen all the items, asking them to rate the system as a whole, on both measures.

### 4.1 Expt.2: Results & Discussion

The results replicate the patterns of results found for correctness and trust (in Expt.1 and KQK) and, again, showed few/no effects for the reasonableness/satisfaction measures. However, the overall ratings of the system's correctness and reasonableness provide converging supporting results.

**4.1.1 Correctness & Reasonableness.**  The MANOVA analyses of correctness ratings revealed significant effects for Explanation and Percent-Error (see Table 1). The interaction was not statistically significant. As before, all of these effects occurred in people's ratings of wrong classifications, not of right classifications.  Univariate main effects for the wrong classifications were found for the Explanation $F(1, 176)=8.24$, $p < .01$, $\eta^2 = 0.05$, and Percent-Error variables, $F(3, 176) = 7.11$, $p < .001$, $\eta^2 = 0.11$. But, no significant effects were found for these variables on right

Table 1. The MANOVA Analyses for Experiments 1 and 2 on the Measures used for the Percent-Error and Explanation Variables

|  |  | Percent-Error | | | | Explanation | | | |
| --- | --- | --- | --- | --- | --- | --- | --- | --- | --- |
|  |  | Wilks' Λ | df | F | $\eta^2$ | Wilks' Λ | df | F | $\eta^2$ |
| Expt.1 | Correctness | 0.707 | (4, 316) | 14.976*** | 0.16 | 0.96 | (2, 158) | 3.129* | 0.038 |
|  | Reasonableness | 0.958 | (4, 316) | 1.715 | 0.021 | 0.975 | (2, 158) | 2.029 | 0.025 |
|  | Trust | 0.616 | (16, 306) | 5.234*** | 0.215 | 0.97 | (8, 153) | 0.598 | 0.03 |
|  | Satisfaction | 0.838 | (16, 304) | 1.755* | 0.085 | 0.954 | (8, 152) | 0.911 | 0.046 |
| Expt.2 | Correctness | 0.879 | (6, 350) | 3.897*** | 0.063 | 0.955 | (2, 175) | 4.12* | 0.045 |
|  | Reasonableness | 0.957 | (6, 350) | 1.29 | 0.022 | 0.986 | (2, 175) | 1.285 | 0.014 |
|  | Global Corr/Reas | 0.74 | (6, 348) | 9.425*** | 0.14 | 0.991 | (2, 174) | 0.762 | 0.009 |
|  | Trust | 0.778 | (24, 490.753) | 1.845*** | 0.08 | 0.934 | (8, 169) | 1.494 | 0.066 |
|  | Satisfaction | 0.902 | (24, 490.753) | 0.739 | 0.034 | 0.919 | (8, 169) | 1.871 | 0.081 |

*$p < 0.05$; **$p < 0.01$; ***$p < .001$

classifications (all $ps > .30$). For example, overall people in the Explanation-present condition rated wrong classifications as more correct ($M = 1.40$, $SD = 0.41$) than those in the Explanation-absent condition ($M = 1.24$, $SD = 0.37$), but both conditions were almost identical for right classifications. Similarly, the Percent-Error variable's effects all occur in wrong classifications, $F(3, 176) = 7.11$, $p < .001$, $\eta^2 = 0.11$, not right classifications, $F(3, 176) = 1.23$, $p > .30$, $\eta^2 = 0.02$. Indeed, this analysis echoed the counterintuitive result from Expt. 1; namely, that the more errors people see, the more they rated them as being correct. Post-hoc Tukey HSD comparisons revealed a significant 0.31 mean increase ($p = .001$) from the 4% to the 20%-error conditions, as well as a 0.27 mean increase ($p < .05$) from the 4% to the 28%-error conditions. There was also a significant 0.27 mean increase ($p < .01$) from the 12%-error to the 20%-error condition (see Fig. 2). MANOVA analyses of reasonableness revealed no reliable effects.

**4.1.2 Global Correctness and Reasonableness.** The MANOVA analyses of global correctness and reasonableness ratings revealed significant effects for Percent-Error, $F(6, 348) = 9.43$, $p < .001$, Wilks' Λ = 0.74, $\eta^2 = 0.140$. No statistically significant effects were found for Explanation or the interaction. The Percent-Error effect was significant both for the global correctness score, $F(3, 175) = 15.34$, $p < .001$, $\eta^2 = 0.21$; as well as for the global reasonableness score, $F(3, 175) = 10.871$, $p < .001$, $\eta^2 = .16$. However, most of this effect can be attributed to the 4%-error condition. Post-hoc Tukey HSD comparisons revealed a higher global correctness and reasonableness ratings for the overall system for the 4%-error condition versus all others. So, this global measure of correctness (like the trust measure) shows that the local effects for Explanation do not persist into a global change in correctness. The Percent-Errors variable was also found to impact overall global correctness and reasonableness, showing people are sensitive to changes in error-rates from 4% to 12% and 28%.

**4.1.3 Trust and Satisfaction.** The MANOVA analyses of global trust ratings of the overall system also revealed a significant effect but no effects for Explanation or an interaction (see Table 1). Univariate main effects for Percent-Error were found in for half of the 8 questions (1, 2, 4, 6). Overall people in the 4%-error condition reported the highest level of confidence in the system ($M = 4.43$, $SD = 0.54$), found the system to be more reliable ($M = 3.24$, $SD = 1.02$) and reported trusting the system to make accurate classifications ($M = 3.93$, $SD = 0.8$) than for other Percent-Errors. In contrast, the MANOVA analyses of satisfaction ratings show no main effects or interaction.

## 5 Related Work

This work sits at the intersection of people's use of automated systems (see [Burton *et al*., 2019]) with an XAI twist. As such, the relevant related work is on (i) trust and error-rates in automated systems, (ii) post-hoc case-base explanations and (iii) XAI work using the MNIST dataset.

**Trust in Automated Systems**. There is a significant literature on what impacts people's trust in automated systems [Burton, *et al.,* 2019; Hoffmann *et al*., 2013]. deVries *et al.* [2003] showed that people's experience of low (20%) versus high (60%) error-rates in a route planner significantly impacted trust; for related work see [Ignaki *et al.*, 1998; Lee and Moray, 1992; Merritt and Ilgen, 2008]. [Dzindolet *et al.,* 2003] presented people with items from a pseudo-computer program that supposedly detected camouflaged soldiers in 200 pictures finding that, when people thought the program was making errors, they immediately started to distrust it, unless they received explanations as to why those errors arose. Dietvorst *et al.*'s [2015] work on algorithm aversion has shown that any forecasting error made by an automated system leads people to prefer their own or other people's forecasts, even when the automated system is demonstrably better than humans. Ribeiro *et al.* [2016] reported a user study with ML-graduate students (N=27) showing a lowering of trust for a "bad classifier" that made 2 incorrect classifications out of 10. The present findings are consistent with this existing literature and suggest that trust in automated systems generalizes over diverse classifiers. The present study shows trust is undermined after seeing a few errors in one continuous interaction with an AI system.

**Post-Hoc Case-Based Explanation User-Studies.** There is a significant literature on post-hoc case-based explanations in CBR; for reviews, see [Sørmo et al., 2005; Leake and McSherry, 2005; Keane and Kenny, 2019a, 2019b] and recommender systems [Tintarev and Masthoff, 2007; Nunes and Jannach, 2017] that contextualizes the current work, though a minority carry out user studies. Nugent and Cunningham [2005; Cunningham *et al.*, 2003], in an exception to this generalization, reported a small user-study on the use of case-based explanations in a blood-alcohol-prediction domain. More recently, an excellent user study by Dodge *et al.* [2019] tested for the effects of four distinct post-hoc explanation strategies on a user's global and local fairness evaluations of a machine learning model. They showed that counterfactual-explanation strategies did best and that the case-based strategy lagged. However, they summarized case information statistically (e.g., "the training set contained 10 individuals identical to X, 60% of these re-offended"), rather than presenting specific individual cases (as is done in most studies). [Yang *et al.*, 2020] tested users (N=33) trust in example-based explanations of a classifier's predictions for images of tree-leaves, finding that specific visual representations improved trust in the system (specifically, "appropriate trust"); their classifier had an accuracy of 71% but, notably, their participants were perhaps less expert (i.e., not botanists) and trust was assessed item-by-item. Finally, [Buçinca *et al.*, 2020] reported two experiments involving the influence of case-based explanations on an AI-model making predictions about fatty-ingredients from pictures of food-dishes; they provided explanations in two different modes, based on multiple cases (four photos of similar food dishes) and a single case with highlighted features (photo of one food-dish with identified ingredients). They found that the provision of explanations improved performance on the fat-estimating task and that the different modes had different effects (n.b., query on veracity of subjective trust/satisfaction measures). This study shows that one can quantifiably assess subjective measures of the human end-user experience.

**MNIST User Studies**. A small number of user-studies have been reported on classification-debugging tasks using the MNIST dataset. Bäuerle *at al.* [2018] built an interface to present miss-classifications to users, grouping them together to speed-up the judgement task to aid model debugging, but their user study was really just a pilot (N=9). The XAI DARPA program reports several groups (notable Rutgers University) that have carried out initial user evaluations of AI classifiers using MNIST [Glickenhaus *et al.*, 2019]; however, this paper was a preliminary report with few details, though notably, they also report that explanations tend only to affect errors, not correct items. Ross and Doshi-Velez (2018) had users (N=11) make plausibility and reasonableness judgements about the robustness of different deep learning methods to adversarial attacks involving the MNIST dataset; though this study was not about explanation or trust *per se*.

## 6 General Discussion

The present experiments reveal a number of novel findings on the use of post-hoc, case-based explanations for a black-box classifier, implemented in the twin-systems approach to XAI (Kenny and Keane, 2019). The pattern of findings represent important results on how (i) local explanations work to impact perceptions the correctness of miss-classifications by the model, (ii) error rates impact perceptions of overall correctness, reasonableness and trust in the model. These results reveal people's tolerance/intolerance for error-rates in AI models and how explanations impact those responses. To summarize:

- *Case-based explanations* impact people's correctness assessments of miss-classifications by the model, locally modifying people's mental models for system predictions (reflected in correctness ratings)
- *Error-rates* impact people's local correctness perceptions of the model's miss-classifications and global measures of correctness, reasonableness and trust; also, people seem quite sensitive to changes in error-levels at the low-end of the scale (e.g., 4% versus 12% errors), though this finding must be viewed in the light of people's high-expertise in judging handwritten numbers (i.e., as experts they may not tolerate errors)
- *Reasonableness* and *Satisfaction* measures, perhaps, require further attention as to their robustness and what they may be actually measuring; both seem to engender very different interpretations in the user population.

With respect to future research the findings of the current study highlight an avenue for additional future research on the influence of error-rates on the trust of an overall system. The experiments described herein used a dataset that contained materials that are familiar to our participants (i.e., they are domain experts). Future research focusing on domains in which users have less knowledge may further challenge the perception of system and explanation validity. In conclusion, the present studies present a rich set of findings for wider consideration of the dynamics of user interactions with explanation-strategies and error-rates in the XAI context. This work also adds significantly, to the (now) small pool of carefully controlled user studies on how explanations and model performance impact AI systems now and in the future.

### Acknowledgements

This paper emanated from research funded by (i) Science Foundation Ireland (SFI) to the *Insight Centre for Data Analytics* (12/RC/2289_P2), (ii) SFI and DAFM on behalf of the Government of Ireland to the *VistaMilk SFI Research Centre* (16/RC/3835), and (iii) the *SFI Centre for Research Training in Machine Learning* (18/CRT/6183).

# References

[Aamodt and Plaze, 1994] Aamodt, A. and Plaza, E.: Case-based reasoning: Foundational issues, methodological variations, and system approaches. AI communications, 7(1), 39-59 1994.

[Bäuerle et al. 2018] Bäuerle, A., Neumann, H., & Ropinski, T.: Training De-Confusion: An Interactive, Network-Supported Visual Analysis System for Resolving Errors in Image Classification Training Data. arXiv preprint arXiv:1808.03114. 2018.

[Buçinca et al., 2020] Buçinca, Z., Lin, P., Gajos, K. Z., & Glassman, E. L.: Proxy tasks and subjective measures can be misleading in evaluating explainable AI systems. In Proceedings of the 25th International Conference on Intelligent User Interfaces pp. 454-464 2020.

[Burton et al., 2019] Burton, J. W., Stein, M. K., & Jensen, T. B.: A systematic review of algorithm aversion in augmented decision making. Journal of Behavioral Decision Making. 2019.

[Cunningham et al., 2003] Cunningham, P., Doyle, D. and Loughrey, J.: An evaluation of the usefulness of case-based explanation. In Case-Based Reasoning Research and Development, vol. 2689, K. D. Ashley and D. G. Bridge, Eds. Berlin, Heidelberg: Springer Berlin Heidelberg, pp. 122–130 2003.

[de Vries et al., 2003] de Vries, P., Midden, C., & Bouwhuis, D.: The effects of errors on system trust, self-confidence, and the allocation of control in route planning. International Journal of Human-Computer Studies, 58(6), 719-735 2003.

[Dietvorst et al., 2015] Dietvorst, B.J., Simmons, J.P. and Massey, C.: Algorithm aversion: People erroneously avoid algorithms after seeing them err. Journal of Experimental Psychology: General, 144, p.114 2015.

[Dodge et al., 2019] Dodge, J., Liao, Q.V., Zhang, Y., Bellamy, R.K. and Dugan, C.: Explaining models: an empirical study of how explanations impact fairness judgment. In Proceedings of the 24th International Conference on Intelligent User Interfaces. pp. 275-285 2019.

[Doshi-Velez and Kim, 2017] Doshi-Velez, F. and Kim, B. Towards a rigorous science of interpretable machine learning. arXiv preprint arXiv:1702.08608, 2017.

[Dzindolet et al., 2003] Dzindolet, M. T., Peterson, S. A., Pomranky, R. A., Pierce, L. G., & Beck, H. P.: The role of trust in automation reliance. International journal of human-computer studies, 58, 697-718 2003.

[Erhan et al., 2009] Erhan, D., Bengio, Y., Courville, A., Vincent, P.:Visualizing higher-layer features of a deep network. University of Montreal, 1341, 1, 2009.

[Giplin et al., 2019] Gilpin, L. H., Testart, C., Fruchter, N., & Adebayo, J.: Explaining explanations to society. arXiv preprint arXiv:1901.06560, 2019.

[Gunning, 2017] Gunning, D. Explainable artificial intelligence (XAI), Defense Advanced Research Projects Agency (DARPA), Web 2. 2017.

[Gunning and Aha, 2019] D. Gunning, D. W. and Aha, DARPA's explainable artificial intelligence program, AI Magazine 40 (2) 2019) 44–58, 2019.

[Glickenhaus et al., 2019] Glickenhaus, B, Karneeb, J. & Aha, D.: DARPA XAI Phase 1 Evaluations Report. DARPA Program, 2019.

[Hoffmann, 2020] Hoffman, R.R.: Personal communication to M.T. Keane, April 2020.

[Hoffmann et al, 2018] R. R. Hoffman, S. T. Mueller, G. Klein, J. Litman: Metrics for explainable ai: Challenges and prospects, arXiv preprint arXiv:1812.04608. 2018.

[Hoffmann et al., 2013] Hoffman, R. R., Johnson, M., Bradshaw, J. M., and Underbrink, A.: Trust in automation. IEEE Intelligent Systems, 28(1), 84-88. 2013.

[Inagaki et al., 1998] Inagaki, T., Moray, N., & Itoh, M.: Trust, self-confidence and authority in human-machine systems. IFAC Proceedings Volumes, 31(26), 431-436 1998.

[Keane and Kenny, 2019a] Keane, M.T. and Kenny, E.M.: The Twin-System Approach as One Generic Solution for XAI. In: IJCAI 2019 Workshop on Explainable Artificial Intelligence (XAI). Macau, China, August 2019.

[Keane and Kenny, 2019b] Keane, M. T. and Kenny, E.M.: How case-based reasoning explains neural networks. In Proceedings of the 27th International Conference on Case Based Reasoning, Sept. 2019.

[Kenny and Keane, 2019] Kenny, E.M. and Keane, M. T.: Twin-systems for explaining ANNs using CBR: Experiments using feature-weighting schemes to retrieve explanatory cases. Proceedings of the 28th International Joint Conference on Artificial Intelligence, August 2019.

[Kenny, Ford, Quinn & Keane, 2020] Kenny, E.M., Quinn, M. & Keane, M. T.: Post Hoc Explanations for Deep Learning: Technique & User Studies. Under review. August. 2020.

[Leake, 1996] Leake, D.B.: CBR in context: The present and future. Case-based reasoning: Experiences, lessons, and future directions, pp.3-30 1996.

[Leake and McSherry, 2005] Leake, D. and McSherry, D.: Introduction to the special issue on explanation in case-based reasoning. Artificial Intelligence Review, 24(2), 103-108 2005.

LeCun, Y., Bottou, L., Bengio, Y., & Haffner, P.: Gradient-based learning applied to document

recognition. Proceedings of the IEEE, 86(11), 2278-2324 1998.

[Lee & Moray, 1992] Lee, J. and Moray, N.: Trust, control strategies and allocation of function in human-machine systems. Ergonomics, 35(10), 1243-1270 1992.

[Li et al., 2017] Li, O., Liu, H., Chen, C. and Rudin, C.: Deep learning for case-based reasoning through prototypes: A neural network that explains its predictions, arXiv:1710.04806 [cs, stat], Oct. 2017.

[Lipton, 2018] Lipton, Z. C.: The mythos of model interpretability. Queue, 16(3), 30 2018.

[Mantaras et al., 2006] De Mantaras, R.L., McSherry, D., Bridge, D., Leake, D., Smyth, B., Craw, S., Faltings, B., Maher, M.L., Cox, M.T., Forbus, K. & Keane, M.T.: Retrieval, reuse, revise and retention in CBR. Knowledge Engineering Review, 20(3), 215-240 2006.

[Merritt & Ilgen, 2008] Merritt, S. M., & Ilgen, D. R.: Not all trust is created equal: Dispositional and history-based trust in human-automation interactions. Human Factors, 50(2), 194-210 2008.

[Miller et al. 2017] Miller, T., Howe, P., & Sonenberg, L.: Explainable AI: Beware of inmates running the asylum or: How I learnt to stop worrying and love the social and behavioural sciences. arXiv preprint arXiv:1712.00547 2017.

[Mueller et al., 2019] Mueller, S. T., Hoffman, R. R., Clancey, W. J., Emery, A. K., & Klein, G.: Explanation in Human-AI Systems: A Literature Meta-Review Synopsis of Key Ideas and Publications and Bibliography for Explainable AI. Florida Institute for Human and Machine Cognition Pensacola United States, 2019.

[Nugent & Cunningham, 2005] Nugent C. and Cunningham, P.: A case-based explanation system for black-box systems, Artificial Intelligence Review, vol. 24, no. 2, pp. 163–178 2005.

[Nunes and Jannach, 2017] Nunes, I. and Jannach, D. A.: systematic review and taxonomy of explanations in decision support and recommender systems. UMUAI, 27(3-5), 393-444, 2017.

[Pedreschi et al., 2019] Pedreschi, D., Giannotti, F., Guidotti, R., Monreale, A., Ruggieri, S. and Turini, F.: Meaningful explanations of black box AI decision systems. In Proceedings of AAAI-19 2019.

[Ribeiro et al., 2016] Ribeiro, M. T., Singh, S., & Guestrin, C.: Why should I trust you?: Explaining the predictions of any classifier. In Proceedings of the 22nd ACM SIGKDD international conference on knowledge discovery and data mining (pp. 1135-1144). ACM 2016.

[Ross & Doshi-valez, 2018] Ross, A. S., & Doshi-Velez, F.: Improving the adversarial robustness and interpretability of deep neural networks by regularizing their input gradients. In Thirty-second AAAI conference on artificial intelligence. 2018.

[Tintarev and Masthoff, 2007] Tintarev, N. and Masthoff, J.: A survey of explanations in recommender systems. In: IEEE 23rd international conference on data engineering workshop, 801-810. IEEE 2007.

[Sørmo et al., 2005] Sørmo, F., Cassens, J. and Aamodt, A.: Explanation in case-based reasoning– perspectives and goals. Artificial Intelligence Review, 24(2), 109-143 2005.

[Weber et al., 2018] Weber, R.O., Johs, A.J., Li, J. and Huang, K.: Investigating textual case-based XAI. In International Conference on Case-Based Reasoning, 431-447. Springer, July, 2018.

[Yang et al., 2020] Yang, F., Huang, Z., Scholtz, J., & Arendt, D. L.: How do visual explanations foster end users' appropriate trust in machine learning?. In Proceedings of the 25th International Conference on Intelligent User Interfaces (pp. 189-201), 2020.